\documentclass[12pt,onecolumn]{IEEEtran}
\usepackage{amssymb}
\usepackage{setspace}
\usepackage{varwidth}
\usepackage[rightbars,color]{changebar}
\usepackage{epsfig}
\usepackage{multirow}
\usepackage{booktabs}
\usepackage[linesnumbered,ruled,vlined]{algorithm2e}
\usepackage{array}
\usepackage{rotating}
\usepackage{epsfig,amsfonts,multirow}
\usepackage[section]{placeins}
\usepackage{graphicx,subfigure}
\usepackage{amsmath}
\usepackage{amsfonts}
\usepackage{caption2}

\ifCLASSINFOpdf

\else

\fi

\begin{document}

\title{Bidirectional-Convolutional LSTM Based Spectral-Spatial Feature Learning for Hyperspectral Image Classification}

\author{Qingshan~Liu,
        ~Feng~Zhou,
        ~Renlong~Hang
        and ~Xiaotong~Yuan

\thanks{These authors are with the Jiangsu Key Laboratory of Big Data Analysis Technology, Jiangsu Collaborative Innovation Center on Atmospheric Environment and Equipment Technology,
Nanjing University of Information Science and Technology, Nanjing 210044, China (qsliu$@$nuist.edu.cn; 13057588879@163.com; renlong\_hang@163.com; xtyuan1980@gmail.com).
}
}

\markboth{}%
{Shell \MakeLowercase{\textit{et al.}}: Bare Demo of IEEEtran.cls for IEEE Journals}

\maketitle
\begin{spacing}{2}
\begin{abstract}
This paper proposes a novel deep learning framework named bidirectional-convolutional long short term memory (Bi-CLSTM) network to automatically learn the spectral-spatial feature from hyperspectral images (HSIs). In the network, the issue of spectral feature extraction is considered as a sequence learning problem, and a recurrent connection operator across the spectral domain is used to address it. Meanwhile, inspired from the widely used convolutional neural network (CNN), a convolution operator across the spatial domain is incorporated into the network to extract the spatial feature. Besides, to sufficiently capture the spectral information, a bidirectional recurrent connection is proposed. In the classification phase, the learned features are concatenated into a vector and fed to a softmax classifier via  a fully-connected operator. To validate the effectiveness of the proposed Bi-CLSTM framework, we compare it with several state-of-the-art methods, including the CNN framework, on three widely used HSIs. The obtained results show that Bi-CLSTM can improve the classification performance as compared to other methods.

\end{abstract}

\IEEEpeerreviewmaketitle

\section{Introduction}
\IEEEPARstart{C}{urrent} hyperspectral sensors can acquire images with high spectral and spatial resolutions simultaneously. For example, the Airborne Visible / Infrared Imaging Spectrometer (AVIRIS) sensor covers 224 continuous spectral bands across the electromagnetic spectrum with a spatial resolution of 3.7 meters. Such rich information has been successfully used in various applications such as urban mapping, environmental management, crop analysis and mineral detection. 

For these applications, an essential step is image classification whose purpose is to identify the label of each pixel. Hyperspectral image (HSI) classification is a challenging task. There exist two important issues \cite{Zhao2016Spectral}, \cite{Hang2015Matrix}. The first one is the curse of dimensionality. HSIs usually contain several hundreds of spectral bands. These high-dimensional data with limited numbers of training samples can easily result in the Hughes phenomenon \cite{Hughes1968On}, which means that the classification accuracy starts to decrease when the number of features exceeds a threshold. The other one is the use of spatial information. The improvement of spatial resolutions may increase spectral variations among intra-class pixels while decrease spectral variations among inter-class pixels \cite{Zhang2012On}, \cite{Xu2014Patch}. Thus, only using spectral information is not enough to obtain a satisfying result.

To solve the first issue, a widely used method is to project the original data into a low-dimensional subspace, in which most of the useful information can be preserved. In the existing literatures, large amounts of works have been proposed \cite{Palsson2015Model}, \cite{Kuo2005Nonparametric}, \cite{Chen2005Local}. They can be roughly divided into two categories: unsupervised FE methods and supervised ones. The unsupervised methods attempt to reveal low-dimensional data structures without using any label information of training samples. Typical methods include but are not limited to principal component analysis (PCA) \cite{Palsson2015Model}, neighborhood preserving embedding (NPE) \cite{He2005Neighborhood}, and independent component analysis (ICA) \cite{Villa2011Hyperspectral}. Different from them, the supervised methods take advantages of the label information to learn the discriminative projections \cite{Zhou2015Dimension}. One typical method is linear discriminant analysis (LDA) \cite{Friedman1989Regularized}, \cite{Bandos2009Classification}, which aims to maximize the inter-class distance and minimize the intra-class distance. In \cite{Kuo2005Nonparametric}, a non-parametric weighted FE (NWFE) method was proposed. NWFE extends LDA by integrating nonparametric scatter matrices with training samples around the decision boundary \cite{Kuo2005Nonparametric}. Local Fisher discriminant analysis (LFDA) was proposed in \cite{Sugiyama2007Dimensionality}, which extends the LDA by assigning greater weights to closer connecting samples.

To address the second issue, many works have been proposed to incorporate the spatial information into the spectral information \cite{Fauvel2013Advances}, \cite{Zhang2013Spatial}, \cite{Sun2015Supervised}. This is because the coverage area of one kind of material or one object usually contains more than one pixel. Current spatial-spectral feature fusion methods can be categorized into three classes: feature-level fusion, decision-level fusion, and regularization-level fusion \cite{Hang2015Matrix}. For feature-level fusion, one often extracts the spatial features and the spectral features independently and then concatenate them into a vector \cite{Zhang2012On}, \cite{Fauvel2008Spectral},  \cite{Mura2011Classification}, \cite{Benediktsson2005Classification}. However, the direct concatenation will lead to a high-dimensional feature space. For decision-level fusion, multiple results are first derived using the spatial and spectral information respectively and then combined according to some strategies such as the majority voting strategy \cite{Tarabalka2009Spectral}, \cite{Jimenez2005Integration}, \cite{Tarabalka2010Multiple}. For regularization-level fusion, a regularizer representing the spatial information is incorporated into the original object function. For example, in \cite{Jia2008Managing} and \cite{Jackson2002Adaptive}, Markov random field (MRF) modeling the joint prior probabilities of each pixel and its spatial neighbors was incorporated into the Bayesian classifier as a regularizer. Although this method works well in capturing the spatial information, optimizing the objective function in MRF is time consuming especially on high-resolution data.

Recently, deep learning (DL) has attracted much attention in the field of remote sensing \cite{Zhang2016Deep}, \cite{liu2016adaptive}, \cite{liu2016learning}. The core idea of DL is to automatically learn high-level semantic features from data itself in a hierarchical manner. In \cite{Chen2014Deep} and \cite{Tao2015Unsupervised}, the autoencoder model has been successfully used for HSI classification. In general, the inputs of the autoencoder model is a high-dimensional vector. Thus, to learn the spatial feature from HSIs, an alternative method is flattening a local image patch into a vector and then feeding it into the model. However, this method may destroy the two-dimensional (2D) structure of images, leading to the loss of spatial information. Similar issues can be found in the deep belief network (DBN) \cite{Chen2015Spectral}. To address this issue, convolutional neural network (CNN) based deep models have been popularly used \cite{Zhao2016Spectral}, \cite{Chen2016Deep}, \cite{Zhao2016Learning}. They directly take the original image or the local image patch as network inputs, and use local-connected and weight sharing structure to extract the spatial features from HSIs. In \cite{Zhao2016Spectral}, the authors designed a CNN network with three convolutional layers and one fully-connected layer. Besides, the input of the network is the first principal component of HSIs extracted by PCA. Although the experimental results demonstrate that this model can successfully learn the spatial feature of HSIs, it may fail to extract the spectral features. Recently, a three-dimensional (3D) CNN model was proposed in \cite{Chen2016Deep}. In order to extract the spectral-spatial features from HSIs, the authors consider the 3D image patches as the input of the network. This complex structure will inevitably increase the amount of parameters, easily leading to the overfitting problem with a limited number of training samples.

In this paper, we propose a bidirectional-convolutional long short term memory (Bi-CLSTM) network to address the spectral-spatial feature learning problem. Specifically, we regard all the spectral bands as an image sequence, and model their relationships using a powerful LSTM network \cite{Hochreiter1997Long}. Similar to other fully-connected networks such as autoencoder and DBN, LSTM can not capture the spatial information of HSIs. Inspired by CNNs, we replace the fully-connected operators in the network by convolutional operators, resulting in a convolutional LSTM (CLSTM) network. Thus, CLSTM can simultaneously learn the spectral and spatial features.  Besides, LSTM assumes that previous states affect future sates, while the spectral channels in the sequence are correlated with each other. To address this issue, we further propose a Bi-CLSTM network. During the training process of the Bi-CLSTM network, we adopt two tricks to alleviate the overfitting problem. They are dropout and data augmentation operations.

\begin{figure}[htp]
  \centering
  \includegraphics[scale = 0.5]{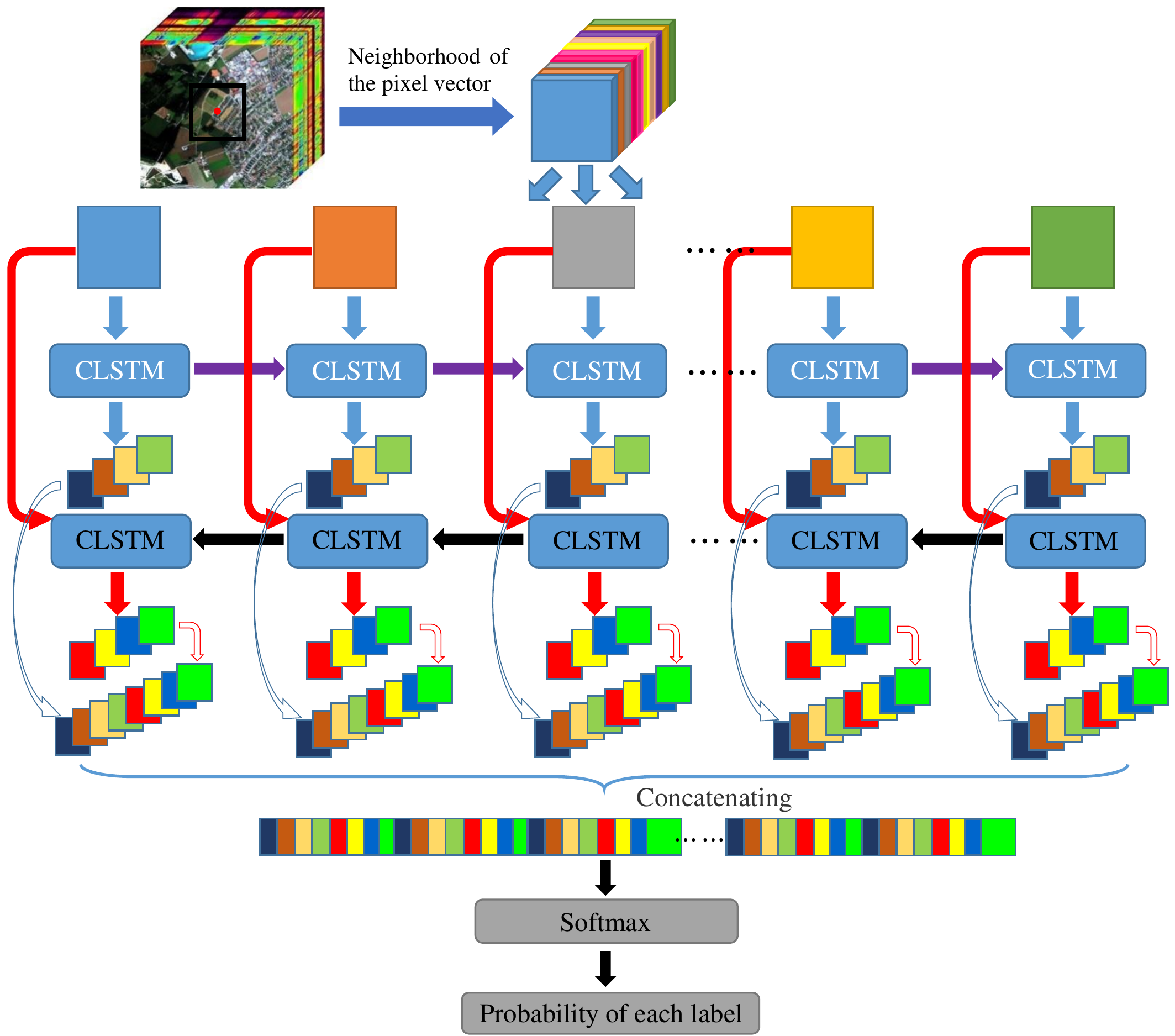}\\
  \caption{Flowchart of the Bi-CLSTM network for HSI classification. For a given pixel, a local cube surrounding it is first extracted, and then unfolded across the spectral domain. The unfolded images are fed into the Bi-CLSTM network one by one.}\label{Flowchart}
\end{figure}
\section{Methodology}
The flowchart of the proposed Bi-CLSTM model is shown in Fig.$~$\ref{Flowchart}. Suppose a HSI can be represented as a 3D matrix $\mathbf{X}\in\mathbf{R}^{{m}\times{n}\times{l}}$ with ${m}\times{n}$ pixels and ${l}$ spectral channels. Given a pixel at the spatial position $({i,j})$ where $1\leq{i}\leq{m}$ and $1\leq{j}\leq{n}$, we can choose a small sub-cube $\mathbf{X}_{ij}\in \mathbf{R}^{{p}\times{p}\times{l}}$ centered at it. The goal of Bi-CLSTM is to learn the most discriminative spectral-spatial information from $\mathbf{X}_{ij}$. Such information is the final feature representation for the pixel at the spatial position $({i,j})$. If we split the sub-cube across the spectral channels, then $\mathbf{X}_{ij}$ can be considered as a ${l}$-length sequence $({x}^{1}_{ij},{x}^{2}_{ij},\cdots,{x}^{l}_{ij})$. The image patches in the sequence are fed into the CLSTM one by one to extract the spectral feature via a recurrent operator and the spatial feature via a convolution operator simultaneously.
\begin{figure}[htp]
  \centering
  \includegraphics[scale = 0.6]{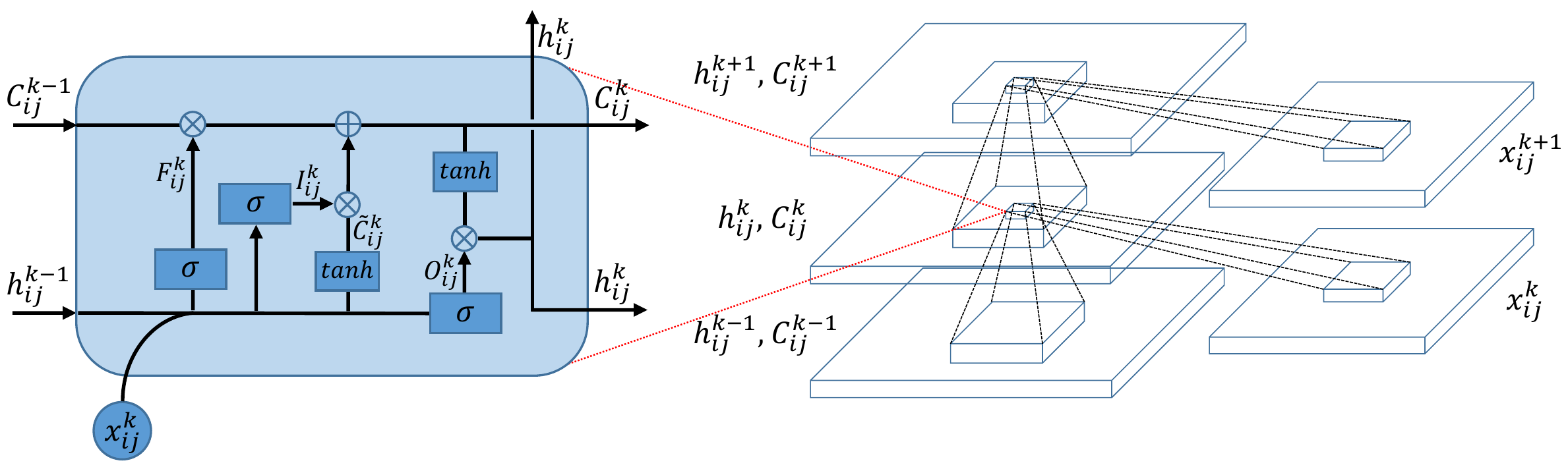}\\
  \caption{The structure of CLSTM.}\label{structure}
\end{figure}

CLSTM is a modification of LSTM. The structure of CLSTM is shown in Fig.$~$\ref{structure}, where the left side zooms in its core computation unit named a memory cell. For the $k$-th image patch ${x}^{k}_{ij}$ in the sequence $\mathbf{X}_{ij}$,
CLSTM firstly decides what information to throw away from the previous cell state ${C}^{k-1}_{ij}$ via the forget gate ${F}^{k}_{ij}$. The forget gate pays attention to ${h}^{k-1}_{ij}$ and ${x}^{k}_{ij}$, and outputs a value between 0 and 1 after an activation function. Here, 1 represents ``keep the whole information" and 0 represents ``throw away the information completely". Secondly, CLSTM needs to decide what new information to store in the current cell state ${C}^{k}_{ij}$. This includes two parts: first, the input gate ${I}^{k}_{ij}$ decides what information to update by the same way as forget gate; second, the memory cell creates a candidate value $\tilde{C}^{k}_{ij}$ computed by ${h}^{k-1}_{ij}$ and ${x}^{k}_{ij}$. After finishing these two parts, CLSTM multiplies the previous memory cell state ${C}^{k-1}_{ij}$ by ${F}^{k}_{ij}$, adds the product to ${I}^{k}_{ij}\circ\tilde{C}^{k}_{ij}$, and updates the information ${C}^{k}_{ij}$. Finally, CLSTM decides what information to output via the cell state ${C}^{k}_{ij}$ and output gate ${O}^{k}_{ij}$. The above process can be formulated as the following equation:
\begin{equation*}\label{CLSTM}
\begin{aligned}
  &{F}^{k}_{ij} = \sigma({W}_{hf}\ast{h}^{k-1}_{ij}+{W}_{xf}\ast{x}^{k}_{ij} + {b}_{f})\\
  &{I}^{k}_{ij} = \sigma({W}_{hi}\ast{h}^{k-1}_{ij}+{W}_{xi}\ast{x}^{k}_{ij} + {b}_{i})\\
  &\tilde{C}^{k}_{ij} = \tanh({W}_{hc}\ast{h}^{k-1}_{ij}+{W}_{xc}\ast{x}^{k}_{ij} + {b}_{c})\\
  &{C}^{k}_{ij} = {F}^{k}_{ij}\circ{C}^{k-1}_{ij} + {I}^{k}_{ij}\circ\tilde{C}^{k}_{ij}\\
  &{O}^{k}_{ij} = \sigma({W}_{ho}\ast{h}^{k-1}_{ij}+{W}_{xo}\ast{x}^{k}_{ij} + {b}_{o})\\
  &{h}^{k}_{ij} = {O}^{k}_{ij}\circ\tanh({C}^{k}_{ij})
\end{aligned}
\end{equation*}
where $\sigma$ is the logistic sigmoid function, `$\ast$' is a convolutional operator, `$\circ$' is a dot product, and ${b}_{f}, {b}_{i}, {b}_{c}$ and ${b}_{o}$ are bias terms. The weight matrix subscripts have the obvious meaning. For example, ${W}_{hi}$ is the hidden-input gate matrix, ${W}_{xo}$ is the input-output gate matrix etc.

In the existing literatures, LSTM has been well acknowledged as a powerful network to address the orderly sequence learning problem based on the assumption that previous states will affect future sates. However,
different from the traditional sequence learning problem, the spectral channels in the sequence are correlated with each other. In \cite{Schuster1997Bidirectional}, bidirectional recurrent neural networks (Bi-RNN) was proposed to use both latter and previous information to model sequential data. Motivated by it, we use a Bi-CLSTM network shown in Fig.$~$\ref{Flowchart} to sufficiently extract the spectral feature. Specifically, the image patches are fed into the CLSTM network one by one with a forward and a backward sequence respectively. After that, we can acquire two spectral-spatial feature sequences. In the classification stage, they are concatenated into a vector and a softmax layer is used to obtain the probability of each class that the pixel belongs to.

It is well known that the performance of DL algorithms depends on the number of training samples. However, there often exists a small number of available samples in HSIs. To this end, we adopt two data augmentation methods. They are flipping and rotating operators. Specifically, we rotate the HSI patches by 90, 180, and 270 degrees anticlockwise and flip them horizontally and vertically. Furthermore, we rotate the horizontally and vertically flipped patches by 90 degrees separately. As a result, the number of training samples can be increased by eight times. Besides the data augmentation method,
dropout \cite{Krizhevsky2012ImageNet} is also used to improve the performance of Bi-CLSTM. We set some outputs of neurons to zeros, which means these neurons do not propagate any information forward or participate in the back-propagation learning algorithm. Every time an input is sampled, network drops neurons randomly to form different structures. In the next section, we will validate the effectiveness of data augmentation and dropout methods.

\section{Experimental Results}
\subsection{Datasets}
\begin{figure}[htp]
  \centering
  \includegraphics[scale = 0.5]{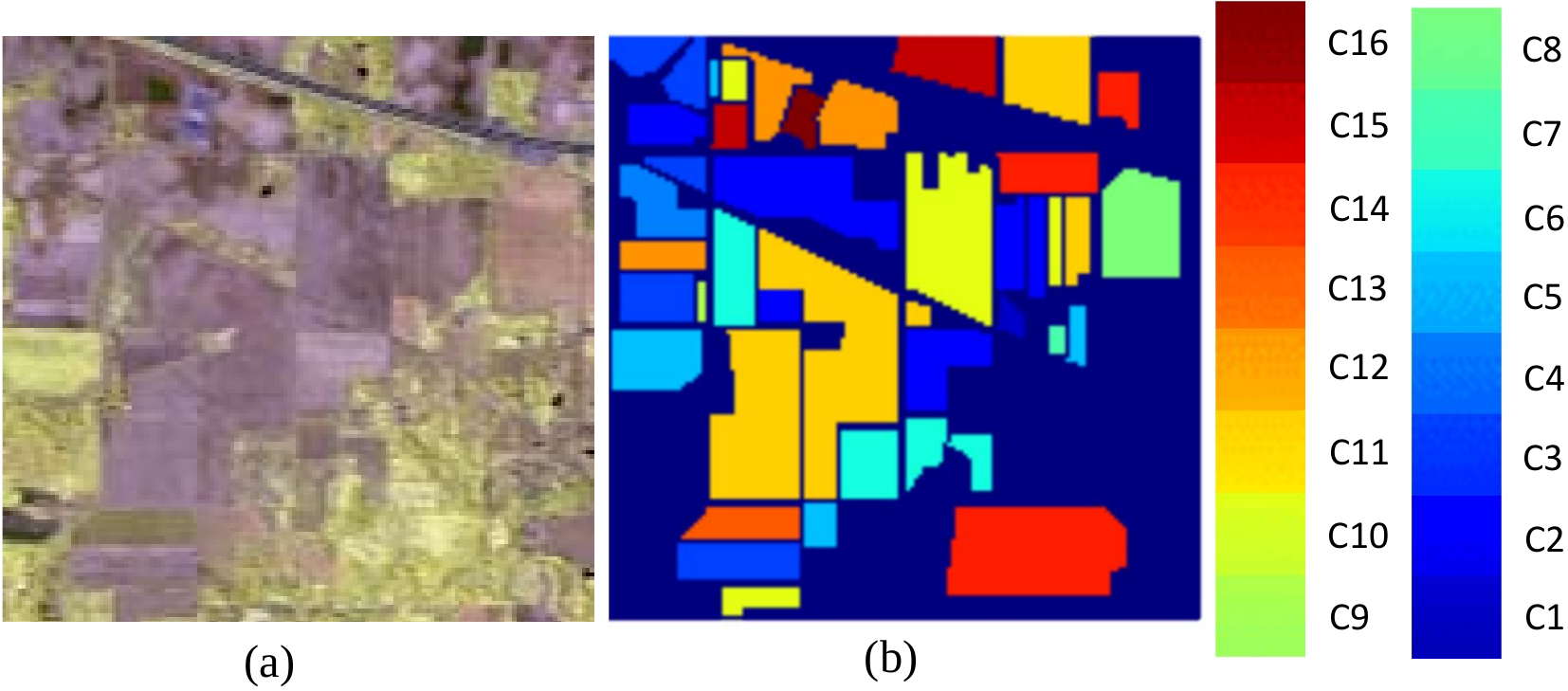}\\
  \caption{Indian Pines scene dataset. (a) False-color composite of the Indian Pines scene. (b) Ground-truth map containing 16 mutually exclusive land cover classes.}\label{IP_show}
\end{figure}

\begin{figure}
  \centering
  \includegraphics[scale = 0.5]{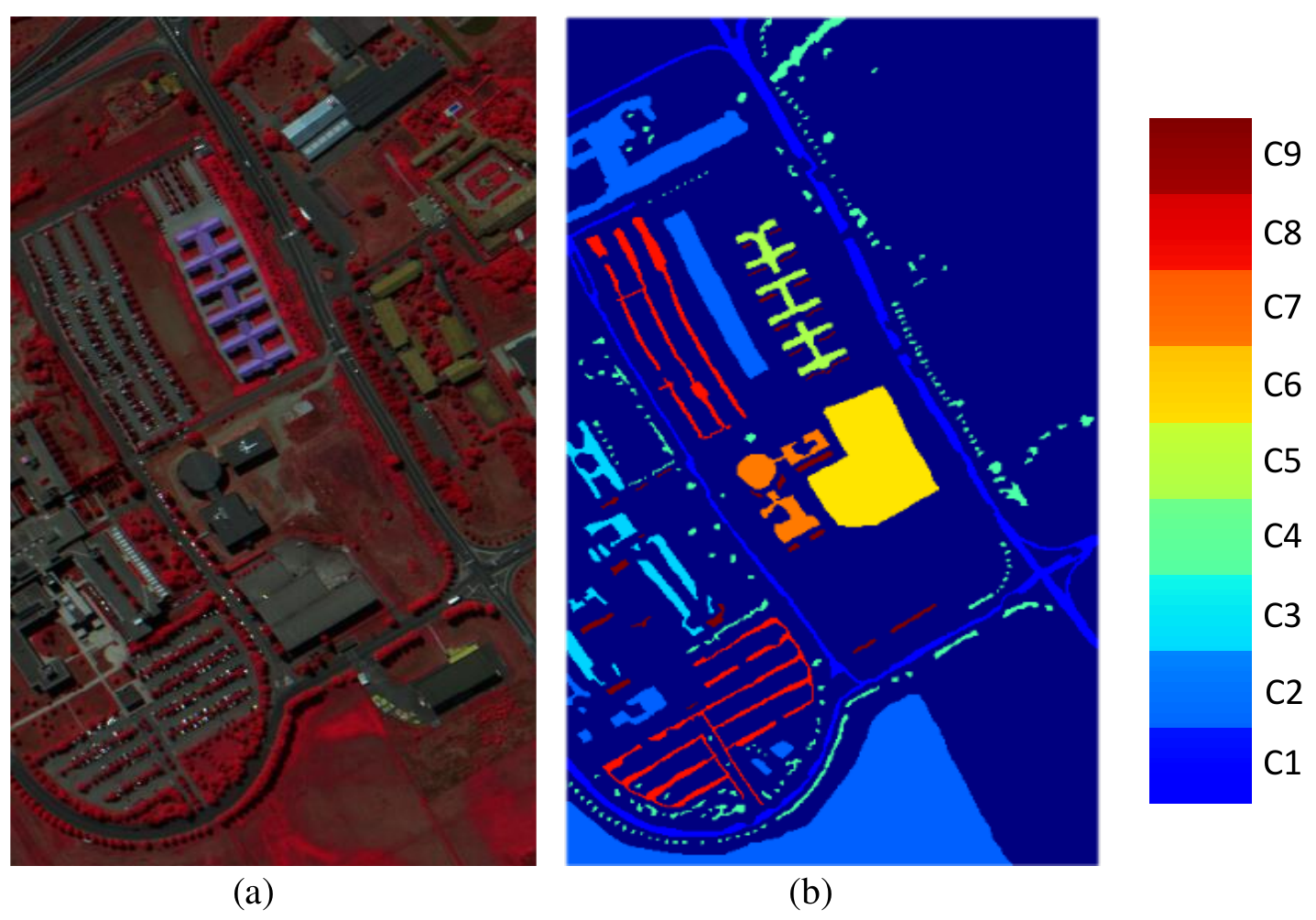}\\
  \caption{Pavia University scene dataset. (a) False-color composite of the Pavia University scene. (b) Ground-truth map containing 9 mutually exclusive land cover classes.}\label{PUS_show}
\end{figure}

\begin{figure}[htp]
  \centering
  \includegraphics[scale = 0.5]{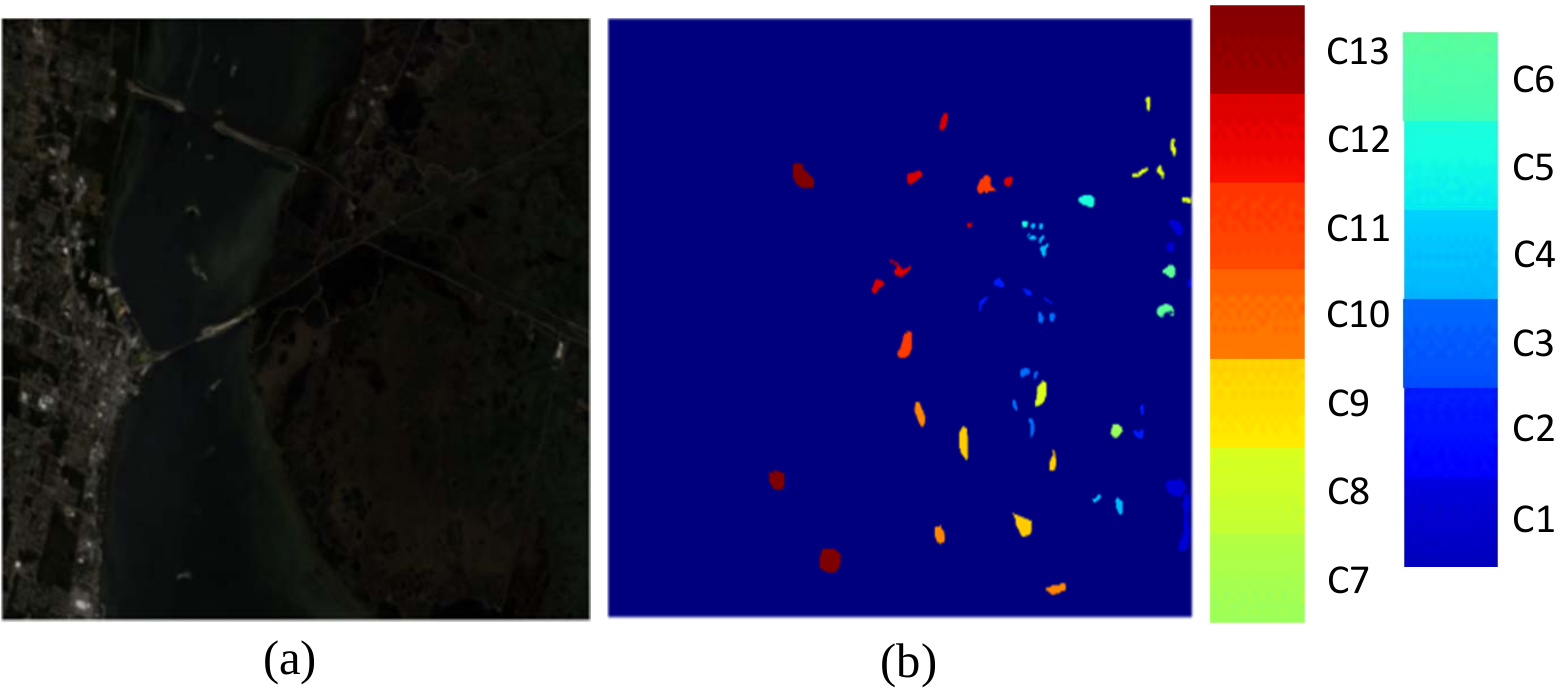}\\
  \caption{KSC dataset. (a) False-color composite of the KSC. (b) Ground-truth map containing 13 mutually exclusive land cover classes.}\label{KSC_show}
\end{figure}

We test the proposed Bi-CLSTM model on three HSIs, which are widely used to evaluate classification algorithms.
\begin{itemize}
\item Indian Pines: The third dataset was acquired by the AVIRIS sensor over the Indian Pine test site in northwestern Indiana, USA, on June 12, 1992 and it contains 224 spectral bands. We utilize 200 bands after removing four bands containing zero values and 20 noisy bands affected by water absorption. The spatial size of the image is $145\times145$ pixels, and the spatial resolution is 20 m. The false-colour composite image and the ground-truth map are shown in Fig.$~$\ref{IP_show}. The available number of samples is 10249 ranging from 20 to 2455 in each class.
  \item Pavia University: The first dataset was acquired by the ROSIS sensor during a flight campaign over Pavia, northern Italy, on July 8, 2002. The original image was recorded with 115 spectral channels ranging from 0.43 $\mu{m}$ to 0.86 $\mu{m}$. After removing noisy bands, 103 bands are used. The image size is $610\times340$ pixels with a spatial resolution of 1.3 m. A three band false-colour composite image and the ground-truth map are shown in Fig.$~$\ref{PUS_show}. In the ground-truth map, there are nine different classes of land covers with more than 1000 labeled pixels for each class.
  \item Kennedy Space Center (KSC): The second dataset was acquired by the AVIRIS sensor over Kennedy Space Center (KSC), Florida, on March 23, 1996. It contains 224 spectral bands. We utilize 176 bands of them after removing bands with water absorption and low signal noise ratio. The spatial size of the image is $512\times614$ pixels, and the spatial resolution is 18 m. Discriminating different land covers in this dataset is difficult due to the similarity of spectral signatures among certain vegetation types. For classification purposes, thirteen classes representing the various land-cover types that occur in this environment are defined. Fig.$~$\ref{KSC_show} demonstrates a false-colour composite image and the ground-truth map.
\end{itemize}

\begin{table}\small
\centering
\caption{Number of pixels for training/testing and the total number of pixels for each class in the Indian Pines ground truth map.}
\label{IPT}
\begin{tabular}{|c|c|c|c|c|c|c|c|c|c|}
\hline
No.  &  Class                & Total       & Training  & Test  & No.  & Class                 & Total                  & Training & Test        \\ \hline
1    & Alfalfa               & 46          &   5       &  41   &  9   & Oats                   &  20                    & 2        &  18         \\\hline
2    & Corn-notill           & 1428        &   143     &  1285 &  10  & Soybean-notill         &  972                    & 97       &  875        \\\hline
3    & Corn-mintill          & 830         &   83      &  747  &  11  & Soybean-mintill        &   2455                  & 246      &  2209       \\\hline
4    & Corn                  & 237         &   24      &  213  &  12  & Soybean-clean           &   593                  & 59       &  534        \\\hline
5    & Grass-pasture         & 483         &   48      &  435  &  13  & Wheat                    &   205                  & 21       &  184        \\\hline
6    & Grass-trees           & 730         &   73      &  657  &  14  & Woods                     &   1265                & 127      &  1138    \\\hline
7    &  Grass-pasture-mowed  & 28          &   3       &  25   &  15  & Buildings-Grass-Trees-Drives&    386              & 39       &  347     \\\hline
8    & Hay-windrowed         & 478         &   48      &  430  &  16  & Stone-Steel-Towers       &    93                  & 9        &  84         \\ \hline
\end{tabular}
\end{table}

\begin{table}\small
\centering
\caption{Number of pixels for training/testing and the total number of pixels for each class in the Pavia University ground truth map.}
\label{PUST}
\begin{tabular}{|c|c|c|c|c|}
\hline
No.  &  Class                     & Total         & Training  & Test       \\\hline
1    & Asphalt                    & 6631           &   548       &  6083   \\\hline
2    & Meadows                    & 18649          &   540     &  18109    \\\hline
3    & Gravel                     & 2099           &   392      &  1707    \\\hline
4    & Trees                      & 3064           &   524      &  2540    \\\hline
5    & Painted metal sheets       & 1345           &   265      &  1080    \\\hline
6    & Bare Soil                  & 5029           &   532      &  4497    \\\hline
7    &  Bitumen                   & 1330           &   375       &  955    \\\hline
8    & Self-Blocking Bricks       & 3682           &   514      &  3168    \\\hline
9    & Shadows                    & 947            &   231      &  716     \\\hline
\end{tabular}
\end{table}

\begin{table}\small
\centering
\caption{Number of pixels for training/testing and the total number of pixels for each class in the KSC ground truth map.}
\label{KSCT}
\begin{tabular}{|c|c|c|c|c|c|c|c|c|c|}
\hline
No.  &  Class                    & Total           & Training  & Test  & No.  & Class             & Total          & Training & Test        \\ \hline
1    & Scrub                      &   761        &   76       &  685   &  8   & Graminoid marsh  &  431      & 43        &  388         \\\hline
2    & Willow swamp               &   243          &   24     &  219 &  9  & Spartina marsh      &  520      & 52       &  468        \\\hline
3    & Cabbage palm hammock       &   256          &   26      &  230  &  10  & Cattail marsh    &  404       & 40      &  364       \\\hline
4    & Cabbage palm/oak hammock   &    252         &   25      &  227  &  11  & Salt marsh       &  419      & 42       &  377        \\\hline
5    & Slash pine                 &   161          &   16      &  145  &  12  & Mud flats        &  503       & 50       &  453        \\\hline
6    & Oak/broadleaf hammock      &   229           &   23      &  206  &  13  & Water           &  927          & 93      &  834       \\\hline
7    &  Hardwood swamp            &   105          &   11       &  94   &      &                &                &           &  \\ \hline
\end{tabular}
\end{table}

\begin{table}\small
\centering
\caption{Detailed configuration of Bi-CLSTM.}
\label{configuration}
\begin{tabular}{|c|c|c|c|c|}
\hline
No.  &  Layer                     & Dropout        & Max-pooling       & Convolution                               \\ \hline
1    & Forward CLSTM              & 0.6            &  2$\times$2       & 3$\times$3$\times$32                      \\ \hline
2    & Backward CLSTM             & 0.6            &  2$\times$2       & 3$\times$3$\times$32                      \\ \hline
\end{tabular}
\end{table}

\subsection{Experimental Setup}
\begin{table}\small
\centering
\caption{OA of Bi-CLSTM with different sizes of input image patches on the KSC dataset.}
\label{Size}
\begin{tabular}{|c|c|c|c|c|}
\hline
Size  &  $8\times8$      &$16\times16$         &$32\times32$      &$64\times64$           \\ \hline
OA(\%)           & 96.12            & 97.78              &98.57	             &99.13                 \\ \hline
\end{tabular}
\end{table}

\begin{table}\small
\centering
\caption{OA of F-CLSTM and Bi-CLSTM on the KSC dataset.}
\label{F_Bi}
\begin{tabular}{|c|c|c|}
\hline
Network  &  F-CLSTM                     & Bi-CLSTM           \\ \hline
OA(\%)    & 95.44                    & 99.13            \\ \hline
\end{tabular}
\end{table}

\begin{table}\small
\centering
\caption{OA of Bi-CLSTM on the KSC dataset with and without dropout and data augmentation.}
\label{dropout}
\begin{tabular}{|c|c|c|}
\hline
Operator          &  Yes                     & No           \\ \hline
Dropout           & 99.13                    & 94.41            \\ \hline
Data augmentation & 99.13                    & 95.07            \\ \hline
\end{tabular}
\end{table}
We compared the proposed Bi-CLSTM model with several FE methods, including PCA, LDA, NWFE, RLDE \cite{Zhou2015Dimension}, MDA \cite{Hang2015Matrix}, and CNN \cite{Chen2016Deep} . Additionally, we also directly use the original pixels as a benchmark. For LDA, the within-class scatter matrix $\mathbf{S}_{W}$ is replaced by $\mathbf{S}_{W}+\varepsilon\mathbf{I}$, where $\varepsilon=10^{-3}$, to alleviate the singular problem. The optimal reduced dimensions for PCA, LDA, NWFE and RLDE are chosen from $[2,30]$. For MDA, the optimal window size is selected from a given set $\{3,5,7,9,11\}$. For Bi-CLSTM, we build a bidirectional network with two CSLTM layers to extract features. Similar to CNN, the convolution operation are followed by max-pooling in Bi-CLSTM, and we empirically set the size of convolution kernel to $3\times3$ and the number of convolution kernel to 32. Without loss of generality, we initialize the state of CLSTM to zeros. The detailed configuration of Bi-CLSTM is listed in Table$~$\ref{configuration}.

For Indian Pines and KSC datasets, we randomly select $10\%$ pixels from each class as the training set, and use the remaining pixels as the testing set. The same as the experiments in \cite{Hang2015Matrix}, we randomly choose 3921 pixels as the training set and the rest of pixels as the testing set for the Pavia University dataset. The detailed numbers of training and testing samples are listed from Table$~$\ref{IPT} to Table$~$\ref{KSCT}.

In order to reduce the effects of random selection, all the algorithms are repeated five times and the average results are reported. The classification performance is evaluated by the overall accuracy (OA), the average accuracy (AA), the per-class accuracy, and the Kappa coefficient $\kappa$. OA defines the
ratio between the number of correctly classified pixels to the total number of pixels in the testing set, AA refers to the average of accuracies in all classes, and $\kappa$ is the percentage of agreement corrected by the number of agreements that would be expected purely by chance.

\subsection{Parameter Selection}
\begin{table}\scriptsize
\centering
\caption{OA, AA, per-class accuracy ($\%$), $\kappa$ and standard deviations after five runs performed by eight methods on the Indain Pines dataset using $10\%$ pixels from each class as the training set.}
\label{IP_r}
\begin{tabular}{|c|c|c|c|c|c|c|c|c|}
\hline
Label & Original        & PCA             & LDA             & NWFE            & RLDE            & MDA             & CNN             & Bi-CLSTM       \\ \hline
OA    & 77.44$\pm$0.71  & 72.58$\pm$0.92  & 76.67$\pm$0.86  & 78.47$\pm$0.66  & 80.97$\pm$0.60  & 92.31$\pm$0.43  & 90.14$\pm$0.78  & 96.78$\pm$0.35  \\ \hline
AA    & 74.94$\pm$0.99  & 70.19$\pm$2.08  & 72.88$\pm$1.10  & 76.08$\pm$1.51  & 80.94$\pm$2.12  & 89.54$\pm$3.08  & 85.66$\pm$3.24  & 94.47$\pm$0.83  \\ \hline
$\kappa$ & 74.32$\pm$0.78& 68.58$\pm$1.10  & 73.27$\pm$0.97  & 75.34$\pm$0.78  & 78.25$\pm$0.70  & 91.21$\pm$0.50  & 88.73$\pm$0.90  & 96.33$\pm$0.40\\ \hline
C1    & 56.96$\pm$10.91 & 59.57$\pm$10.03  & 63.04$\pm$11.50 & 62.17$\pm$8.22 & 64.78$\pm$15.25  & 73.17$\pm$17.92  & 71.22$\pm$15.75 & 93.66$\pm$6.12  \\ \hline
C2    & 79.75$\pm$2.77 & 68.75$\pm$1.22 & 72.04$\pm$1.37  & 76.27$\pm$3.38 & 78.39$\pm$1.34  & 93.48$\pm$1.42     & 90.10$\pm$2.33  & 96.84$\pm$2.05 \\ \hline
C3    & 66.60$\pm$3.03 & 53.95$\pm$2.69 & 57.54$\pm$2.67 & 59.64$\pm$4.32  & 68.10$\pm$2.16 & 84.02$\pm$3.11    & 91.03$\pm$2.73  & 97.22$\pm$2.02 \\ \hline
C4    & 59.24$\pm$7.14 & 55.19$\pm$8.85 & 46.58$\pm$5.93 & 59.83$\pm$7.76  & 70.80$\pm$6.04 & 83.57$\pm$2.23     & 85.73$\pm$5.02 & 96.71$\pm$3.59  \\ \hline
C5    & 90.31$\pm$1.45  & 83.85$\pm$2.36  & 91.76$\pm$0.59  & 88.49$\pm$2.39  & 92.17$\pm$1.97  & 96.69$\pm$1.39  & 83.36$\pm$5.75  & 92.28$\pm$3.82 \\ \hline
C6    & 95.78$\pm$1.64 & 91.23$\pm$2.63 & 94.41$\pm$1.95 & 96.19$\pm$1.56 & 94.90$\pm$2.04 & 99.15$\pm$0.51      & 91.99$\pm$3.25  & 99.39$\pm$0.61 \\ \hline
C7    & 80.00$\pm$7.82  & 82.86$\pm$5.87  & 72.14$\pm$19.13 & 82.14$\pm$7.58 & 85.71$\pm$6.68 & 93.60$\pm$6.07    & 85.60$\pm$12.20 & 92.00$\pm$9.80 \\ \hline
C8    & 97.41$\pm$0.84 & 93.97$\pm$3.11 & 98.74$\pm$0.68  & 99.04$\pm$0.48 & 99.12$\pm$0.95 & 99.91$\pm$0.13      & 97.35$\pm$3.75 & 99.91$\pm$0.21\\ \hline
C9    & 35.00$\pm$10.61  & 34.00$\pm$10.84  & 26.00$\pm$14.32  & 44.00$\pm$8.22  & 73.00$\pm$21.10  & 63.33$\pm$24.72&54.45$\pm$23.70  & 76.67$\pm$21.66 \\ \hline
C10    & 66.32$\pm$3.18 & 64.18$\pm$4.19 & 60.91$\pm$1.32 & 69.18$\pm$3.56 & 69.73$\pm$1.07 & 82.15$\pm$2.23     & 75.38$\pm$8.97  & 95.93$\pm$2.00 \\ \hline
C11    & 70.77$\pm$2.42  & 74.96$\pm$1.77  & 76.45$\pm$1.49 & 77.78$\pm$0.49 & 79.38$\pm$0.56 & 92.76$\pm$1.45   & 94.36$\pm$0.48 & 96.31$\pm$1.46 \\ \hline
C12    & 64.42$\pm$3.92 & 41.72$\pm$5.95 & 67.45$\pm$2.13  & 64.05$\pm$5.37 & 72.28$\pm$3.42 & 91.35$\pm$2.26     & 78.73$\pm$8.00 & 93.33$\pm$3.12\\ \hline
C13    & 95.41$\pm$2.62  & 93.46$\pm$2.50  & 96.00$\pm$2.08  & 97.56$\pm$1.89  & 97.56$\pm$1.38  & 99.13$\pm$0.49  & 95.98$\pm$4.82  & 95.76$\pm$3.72 \\ \hline
C14    & 92.66$\pm$1.77 & 89.45$\pm$1.89 & 93.79$\pm$0.96 & 93.49$\pm$1.15 & 92.36$\pm$0.92 & 98.22$\pm$0.39     & 96.80$\pm$1.08  & 99.49$\pm$0.35 \\ \hline
C15    & 60.88$\pm$6.27  & 47.77$\pm$6.29  & 65.54$\pm$4.09 & 58.50$\pm$4.70 & 67.10$\pm$6.39 & 87.84$\pm$4.00    & 96.54$\pm$2.54 & 98.67$\pm$1.11 \\ \hline
C16    & 87.53$\pm$1.95 & 88.17$\pm$2.01 & 83.66$\pm$8.85  & 89.03$\pm$2.78 & 89.68$\pm$3.28 & 94.29$\pm$6.43    & 81.90$\pm$17.71 & 87.38$\pm$9.09\\ \hline
\end{tabular}
\end{table}

\begin{figure}[!htb]
  \centering
  \includegraphics[scale = 0.5]{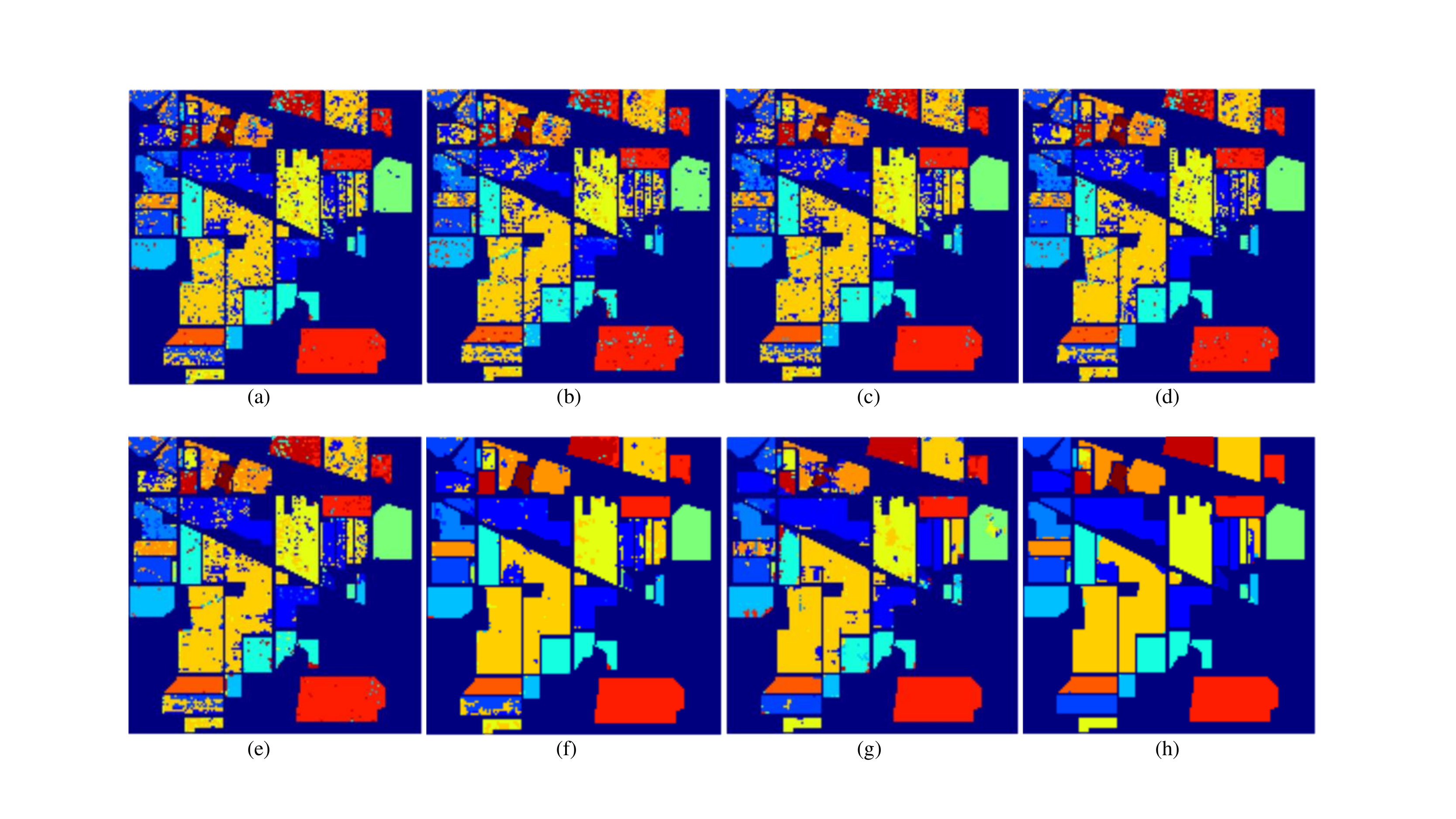}\\
  \caption{Classification maps using eight different methods on the Indian Pines dataset. (a) Original. (b) PCA. (c) LDA. (d) NWFE. (e) RLDE. (f) MDA. (g) CNN. (h) Bi-CLSTM.}\label{IP_rshow}
\end{figure}

\begin{table}\scriptsize
\centering
\caption{OA, AA, per-class accuracy ($\%$), $\kappa$ and standard deviations after five runs performed by eight methods on the Pavia University Scene dataset using 3921 pixels as the training set.}
\label{PUS_r}
\begin{tabular}{|c|c|c|c|c|c|c|c|c|}
\hline
Label & Original        & PCA             & LDA             & NWFE            & RLDE            & MDA             & CNN             & Bi-CLSTM       \\ \hline
OA    & 89.12$\pm$0.26  & 88.63$\pm$0.26  & 84.08$\pm$0.34  & 88.73$\pm$0.27  & 88.82$\pm$0.25  & 96.95$\pm$0.29  & 96.55$\pm$0.85  & 99.10$\pm$0.16  \\ \hline
AA    & 90.50$\pm$0.06  & 90.18$\pm$0.11  & 87.23$\pm$0.18  & 90.38$\pm$0.08  & 90.45$\pm$0.06  & 96.86$\pm$0.23  & 97.19$\pm$0.51  & 99.20$\pm$0.17  \\ \hline
$\kappa$& 85.81$\pm$0.32& 85.18$\pm$0.32  & 79.59$\pm$0.42  & 85.31$\pm$0.35  & 85.43$\pm$0.31  & 95.93$\pm$0.52  & 95.30$\pm$1.13  & 98.77$\pm$0.21\\ \hline
C1    & 87.25$\pm$0.57 & 87.07$\pm$0.84  & 82.91$\pm$1.42 & 86.86$\pm$0.89 & 87.20$\pm$0.52  & 96.69$\pm$0.41  & 96.72$\pm$1.48 & 98.56$\pm$0.58  \\ \hline
C2    & 89.10$\pm$0.54 & 88.38$\pm$0.43 & 80.68$\pm$0.42  & 88.50$\pm$0.41 & 88.40$\pm$0.52  & 97.76$\pm$0.47     & 96.31$\pm$1.75  & 99.23$\pm$0.39 \\ \hline
C3    & 81.99$\pm$1.05 & 81.96$\pm$1.03 & 69.21$\pm$1.17 & 82.20$\pm$0.52  & 81.69$\pm$0.80 & 90.69$\pm$1.44    & 97.15$\pm$1.58  & 99.27$\pm$0.47 \\ \hline
C4    & 95.65$\pm$0.59 & 95.14$\pm$0.49 & 95.99$\pm$0.90 & 95.27$\pm$0.48  & 95.79$\pm$0.56 & 98.44$\pm$0.27     & 96.16$\pm$1.29 & 98.21$\pm$0.92  \\ \hline
C5    & 99.76$\pm$0.14  & 99.76$\pm$0.14  & 99.90$\pm$0.07  & 99.81$\pm$0.08  & 99.87$\pm$0.08  & 100.00$\pm$0.00  & 99.81$\pm$0.32  & 99.87$\pm$0.15 \\ \hline
C6    & 88.78$\pm$1.01 & 88.06$\pm$0.63 & 89.53$\pm$0.76 & 88.16$\pm$0.59 & 88.67$\pm$0.67 & 96.26$\pm$0.45      & 94.87$\pm$3.62  & 99.56$\pm$0.29 \\ \hline
C7    & 85.92$\pm$0.93  & 85.32$\pm$1.54  & 81.11$\pm$0.98 & 86.57$\pm$1.55 & 86.06$\pm$1.04 & 97.95$\pm$0.62    & 97.44$\pm$1.68 & 99.75$\pm$0.30 \\ \hline
C8    & 86.14$\pm$1.02 & 86.06$\pm$0.72 & 85.81$\pm$1.20  & 86.13$\pm$0.73 & 86.42$\pm$0.73 & 93.98$\pm$0.97      & 98.23$\pm$0.91 & 99.82$\pm$0.55\\ \hline
C9    & 99.92$\pm$0.05  & 99.92$\pm$0.05  & 99.92$\pm$0.05  & 99.89$\pm$0.00  & 99.94$\pm$0.06 & 100.00$\pm$0.00  &98.04$\pm$0.96  & 99.53$\pm$0.47 \\ \hline
\end{tabular}
\end{table}

\begin{figure}[!htb]
  \centering
  \includegraphics[scale = 0.5]{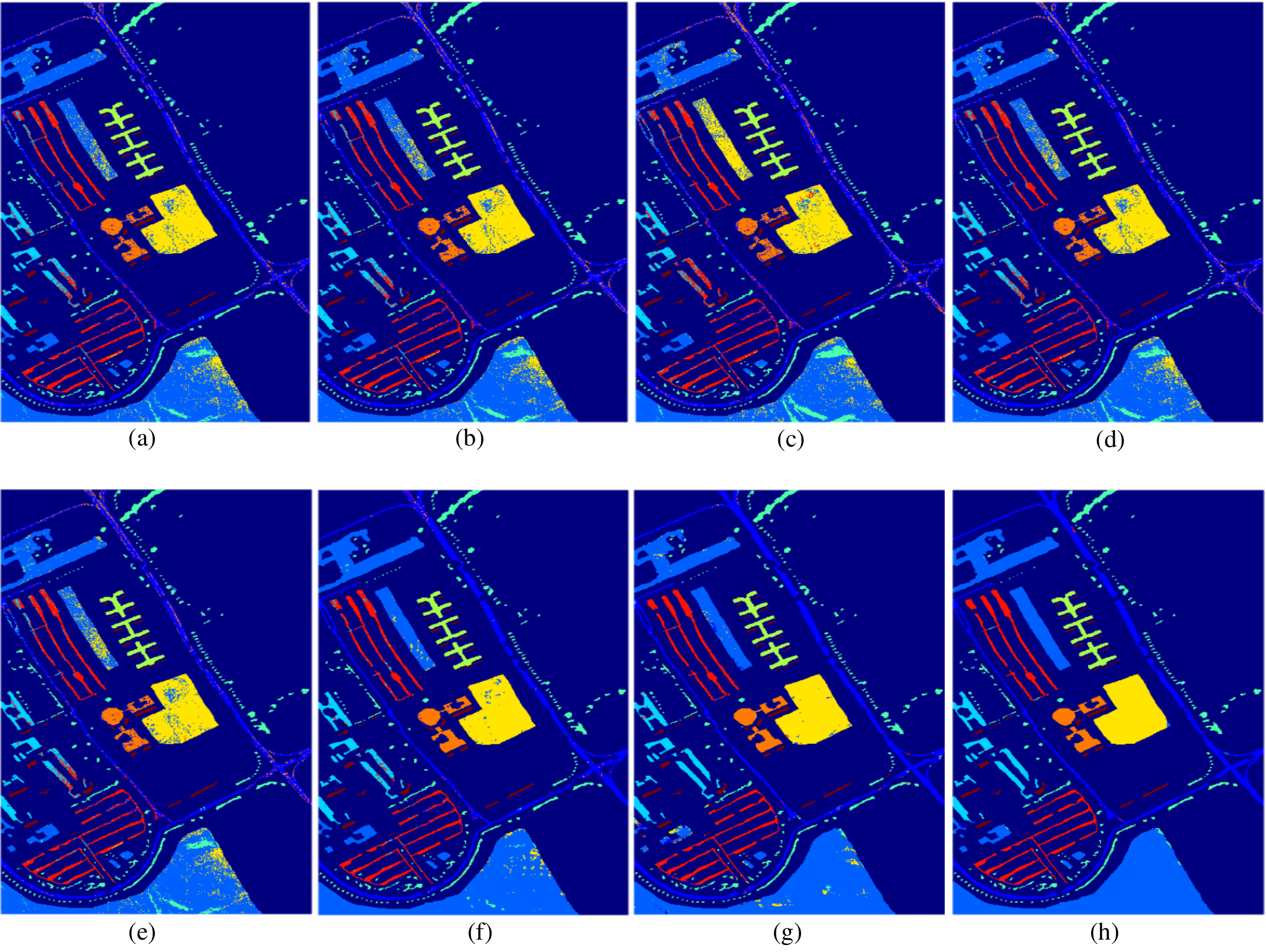}\\
  \caption{Classification maps using eight different methods on the Pavia University Scene dataset. (a) Original. (b) PCA. (c) LDA. (d) NWFE. (e) RLDE. (f) MDA. (g) CNN. (h) Bi-CLSTM.}\label{PUS_rshow}
\end{figure}

\begin{figure}[htb]
  \centering
  \includegraphics[scale = 0.5]{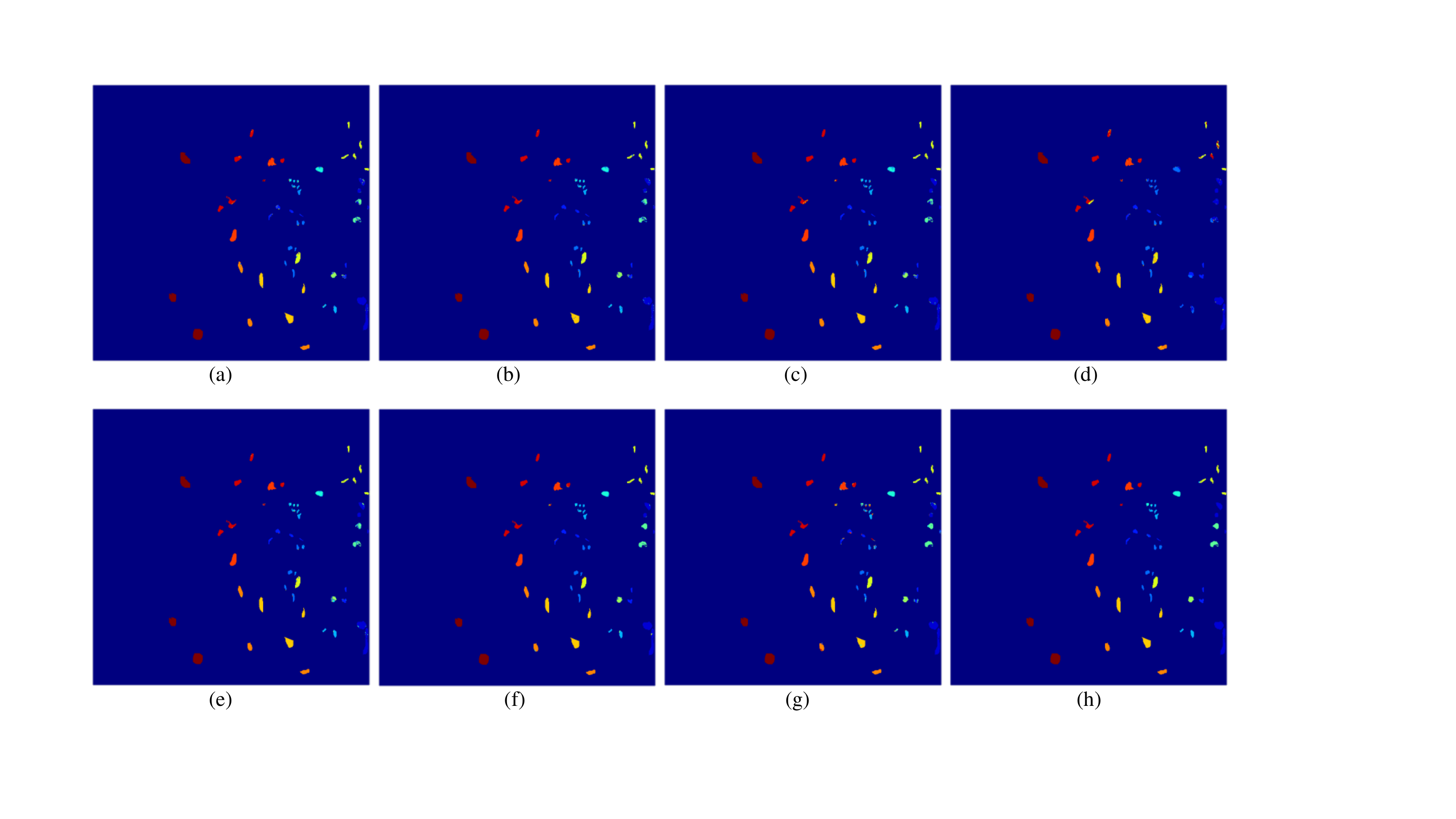}\\
  \caption{Classification maps using eight different methods on the KSC dataset. (a) Original. (b) PCA. (c) LDA. (d) NWFE. (e) RLDE. (f) MDA. (g) CNN. (h) Bi-CLSTM.}\label{KSC_rshow}
\end{figure}

\begin{table}\scriptsize
\centering
\caption{OA, AA, per-class accuracy ($\%$), $\kappa$ and standard deviations after five runs performed by eight methods on the KSC dataset using $10\%$ pixels from each class as the training set.}
\label{KSC_r}
\begin{tabular}{|c|c|c|c|c|c|c|c|c|}
\hline
Label & Original        & PCA             & LDA             & NWFE            & RLDE            & MDA             & CNN             & Bi-CLSTM       \\ \hline
OA    & 93.16$\pm$0.38  & 92.60$\pm$0.58  & 92.05$\pm$0.32  & 75.70$\pm$3.09  & 93.50$\pm$0.31 & 96.81$\pm$0.17  & 92.55$\pm$0.84  & 98.29$\pm$0.98  \\ \hline
AA    & 89.15$\pm$0.55  & 88.45$\pm$0.61  & 87.02$\pm$0.94  & 59.47$\pm$4.42  & 90.09$\pm$0.71 & 95.30$\pm$0.83  & 89.20$\pm$1.50  & 97.77$\pm$1.37  \\ \hline
$\kappa$& 92.38$\pm$0.42 & 91.76$\pm$0.64 & 91.14$\pm$0.36  & 72.65$\pm$3.70  & 92.77$\pm$0.34 & 96.45$\pm$0.18  & 91.69$\pm$0.95  & 98.10$\pm$1.09\\ \hline
C1    & 95.43$\pm$2.54 & 95.14$\pm$2.72  & 95.40$\pm$1.76 & 97.14$\pm$0.69    & 95.30$\pm$1.64 & 96.93$\pm$1.03  & 94.86$\pm$1.30 & 98.87$\pm$1.36  \\ \hline
C2    & 91.44$\pm$4.43 & 91.36$\pm$4.56 & 92.51$\pm$2.32  & 91.19$\pm$5.65    & 92.26$\pm$5.48 & 97.26$\pm$1.29  & 77.53$\pm$5.05  & 93.61$\pm$5.93 \\ \hline
C3    & 90.86$\pm$6.55 & 90.55$\pm$6.05 & 82.89$\pm$3.38 & 77.19$\pm$43.15    & 88.44$\pm$2.00 & 98.92$\pm$0.30  & 84.52$\pm$5.31  & 99.35$\pm$0.56 \\ \hline
C4    & 79.52$\pm$5.74 & 77.94$\pm$6.24 & 71.98$\pm$5.47 & 0.08$\pm$0.18     & 76.90$\pm$5.48  & 90.31$\pm$0.62  & 77.71$\pm$11.85 & 94.71$\pm$2.07  \\ \hline
C5    & 68.20$\pm$7.71 & 65.34$\pm$8.30 & 62.36$\pm$7.59  & 0.00$\pm$0.00    & 77.64$\pm$2.45  & 80.00$\pm$7.80  & 80.97$\pm$9.54  & 97.24$\pm$2.93 \\ \hline
C6    & 67.34$\pm$3.90 & 64.54$\pm$4.25 & 74.93$\pm$4.75 & 4.37$\pm$9.76     & 77.82$\pm$0.72  & 92.47$\pm$2.40  & 72.62$\pm$14.78  & 94.54$\pm$9.01 \\ \hline
C7    & 84.19$\pm$5.33  & 85.52$\pm$6.00 & 72.95$\pm$9.15 & 0.00$\pm$0.00    & 82.67$\pm$16.06 & 94.68$\pm$6.01  & 93.19$\pm$5.35 & 99.74$\pm$0.53 \\ \hline
C8    & 95.17$\pm$1.26 & 94.66$\pm$0.98 & 89.88$\pm$3.41  & 36.33$\pm$16.22  & 91.97$\pm$2.39  & 96.26$\pm$4.19  & 93.87$\pm$2.41 & 97.23$\pm$3.16\\ \hline
C9    & 95.92$\pm$1.69  & 94.15$\pm$2.06 & 95.12$\pm$2.97  & 94.92$\pm$3.62  & 98.08$\pm$1.50  & 99.89$\pm$0.15  & 95.85$\pm$3.03  & 97.81$\pm$1.08 \\ \hline
C10    & 96.78$\pm$1.56 & 96.68$\pm$1.65 & 99.21$\pm$0.64 & 90.10$\pm$2.42   & 96.78$\pm$1.20  & 98.35$\pm$0.39  & 96.81$\pm$1.79  & 99.66$\pm$0.52 \\ \hline
C11    & 98.14$\pm$0.87  & 98.14$\pm$0.87 & 97.85$\pm$0.29 & 94.18$\pm$1.49  & 98.23$\pm$1.38  & 99.33$\pm$0.19  & 94.27$\pm$2.21 & 98.94$\pm$2.12 \\ \hline
C12    & 95.90$\pm$1.23 & 95.83$\pm$1.36 & 96.22$\pm$1.73  & 87.59$\pm$2.59  & 95.39$\pm$1.31  & 94.59$\pm$1.72  & 97.35$\pm$2.09 & 99.28$\pm$0.89\\ \hline
C13    & 100.00$\pm$0.00&100.00$\pm$0.00 &100.00$\pm$0.00  & 99.98$\pm$0.05  & 99.68$\pm$0.40  & 99.94$\pm$0.08  & 100.00$\pm$0.00 & 100.00$\pm$0.00 \\ \hline
\end{tabular}
\end{table}

There are four important influence factors in Bi-CLSTM, including dropout, data augmentation, network framework, and the size of input image patches. Firstly, to find the optimal size of image patches, we
fix the other three factors and select the size from four candidate values $\{8,16,32,64\}$. Table$~$\ref{Size} demonstrates the effects of different sizes on OA of the KSC dataset. From this table, we can observe that OA increases as the patch size increases, and $64\times64$ size can achieve a high enough accuracy. Since larger size will dramatically increase the computation time and the accuracy improvement is limited, the optimal size can be chosen as $64\times64$.

Secondly, to investigate the performance of bidirectional network structure, we fix the other influence factors and compare forward-CLSTM (F-CLSTM) with Bi-CLSTM on the KSC dataset. Here, F-CLSTM is a forward network with the same configuration as Bi-CLSTM listed in Table$~$\ref{configuration}. As shown in Table$~$\ref{F_Bi}, the bidirectional network indeed outperforms the ordinary forward network. This result certifies the effectiveness of Bi-CLSTM as compared to the forward CLSTM.

Finally, we also validate the effectiveness of dropout and data augmentation operators. We set the probability of dropout to the common value 0.6, and fix the other influence factors. Table$~$\ref{dropout} reports the OA values with or without dropout operator on the KSC dataset. It can be observed that using dropout can significantly improve the accuracy from 94.41\% to 99.13\%. Similarly, we expand the number of training samples by eight times as described in Section \uppercase\expandafter{\romannumeral2}-C and fix the other influence factors. Table$~$\ref{dropout} demonstrates that data augmentation can improve the accuracy from 95.07\% to 99.13\%.

\subsection{Performance Comparison}

To demonstrate the superiority of the proposed Bi-CLSTM model, we quantitatively and qualitatively compare it with the aforementioned methods. Table$~$\ref{IP_r} reports the quantitative results acquired by eight methods on the Indian Pines dataset. From these results, we can observe that PCA achieves lowest performance among eight methods, mainly because PCA directly extracts spectral features for classification without considering spatial features. Although LDA and NWFE are still spectral-based FE methods, they achieve better results than PCA due to the use of the label information in the training samples.  Besides, MDA achieves better performance than the other LDA-related methods which consider spectral information only, because it can extract spatial and spectral features simultaneously. This indicates the importance of spatial features for HSI classification. So, as a spatial-based method, CNN performs better than other spectral-based methods. However, CNN only uses the principal component of all spectral bands, leading to the loss of spectral information. Therefore, the performance obtained by CNN is inferior to that by MDA. More importantly, Bi-CLSTM can achieve the highest OA than other methods. Compared to CNN, Bi-LSTM can sufficiently make use of the whole spectral information, thus improving OA from 90.14\% to 96.78\%. Additionally, as a kind of neural network, Bi-CLSTM is able to capture the non-linear distribution of hyperspectral data, while the linear FE method MDA may fail. Therefore, Bi-CLSTM obtains better results than MDA. Fig.$~$\ref{IP_rshow} demonstrates the classification maps achieved by eight different methods on the Indian Pines dataset. It can be observed that Bi-CLSTM obtains more homogeneous maps than other methods.

Similar results are demonstrated in Table$~$\ref{PUS_r} and Fig.$~$\ref{PUS_rshow} on the Pavia University Scene dataset. Again, MDA, CNN, and Bi-CLSTM achieve better performance than other methods. Specifically, OA, AA and $\kappa$ obtained by CNN are almost the same as MDA, and Bi-CLSTM obtains better performance than CNN and MDA. It is worth noting that the improvement of OA, AA and $\kappa$ from MDA or CNN to Bi-CLSTM is not remarkable as those on the Indian Pines dataset, because CNN and MDA have already obtained a high performance and a further improvement is very difficult. Table$~$\ref{KSC_r} and Fig.$~$\ref{KSC_rshow} show the classification results of different methods on the KSC dataset. Similar to the other two datasets, Bi-CLSTM achieves the highest OA, AA and $\kappa$ than other methods.

\section{Conclusion}
In this paper, we propose a novel bidirectional-convolutional long short term memory (Bi-CLSTM) network to automatically learn the spectral-spatial feature from hyperspectral images (HSIs). The input of  the network is the whole spectral channels of HSIs, and a bidirectional recurrent connection operator across them is used to sufficiently explore the spectral information. Besides, motivated by the widely used convolutional neural network (CNN), fully-connected operators in the network is replaced by convolution operators across the spatial domain to capture the spatial information. By conducting experiments on
three HSIs collected by different instruments (AVIRIS and ROSIS), we compare the proposed method with several feature extraction methods including CNN. The experimental results indicate that using spatial
information improves the classification performance and results in more homogeneous regions in classification maps compared to only using spectral information. In addition, the proposed
method can improve the OA, AA, and $\kappa$ on three HSIs as compared to CNN. We also evaluate the influences of different components in the network, including dropout, data augmentation and patch size.

\bibliography{IEEEfull,CLSTM}
\bibliographystyle{IEEEtran}
\end{spacing}
\end{document}